\documentclass[runningheads]{llncs}
\usepackage{graphicx}

\usepackage{tikz}
\usepackage{comment}
\usepackage{amsmath,amssymb} 
\usepackage{color}

\usepackage[accsupp]{axessibility}  

\usepackage[pagebackref,breaklinks,colorlinks]{hyperref}

\usepackage{bm}
\usepackage{caption}
\usepackage{subcaption}
\usepackage{booktabs}
\usepackage{bold-extra}
\usepackage{xspace}
\usepackage{dsfont}
\usepackage{colortbl}
\usepackage[frozencache]{minted}
\usepackage{graphics}
\usepackage{textcomp}
\usepackage{multirow}
\usepackage{algorithm}
\usepackage{algorithmic}

\definecolor{mygray}{gray}{.9}
\newlength{\twosubht}
\newsavebox{\twosubbox}
\newcommand{\mrtwo}[1]{\multirow{2}{*}{#1}}

\newcommand\pointscatter{\texttt{PointScatter}\xspace}
\newcommand\ps{\texttt{PointScatter}\xspace}
\newcommand\psaux{\texttt{PSAUX}\xspace}
\newcommand{\bd}[1]{\textbf{#1}}
\newcommand{\mr}[2]{\multirow{#1}{*}{#2}}

\makeatletter
\DeclareRobustCommand\onedot{\futurelet\@let@token\@onedot}
\def\@onedot{\ifx\@let@token.\else.\null\fi\xspace}

\def\eg{\emph{e.g}\onedot} 
\def\ie{\emph{i.e}\onedot} 
 
\def\etc{\emph{etc}\onedot} 
 
\def\etal{\emph{et al}\onedot}
\makeatother

\makeatletter
\newcommand{\printfnsymbol}[1]{%
  \textsuperscript{\@fnsymbol{#1}}%
}
\makeatother

\PassOptionsToPackage{unicode}{hyperref}
\PassOptionsToPackage{naturalnames}{hyperref}

\begin{document}

\pagestyle{headings}
\mainmatter
\def\ECCVSubNumber{213}  

\title{\texttt{\textbf{PointScatter}}: Point Set Representation for \\ Tubular Structure Extraction}

\titlerunning{Point Set Representation for Tubular Structure Extraction}

\author{Dong Wang\thanks{Equal contribution.}\inst{1} \and
        Zhao Zhang$^\star$\inst{2} \and
        Ziwei Zhao\inst{2,4,5} \and \\
        Yuhang Liu\inst{3} \and
        Yihong Chen\inst{2} \and
        Liwei Wang\inst{1,2}}
\authorrunning{D. Wang et al.}
%
\institute{
{\small Key Laboratory of Machine Perception, MOE, School of Artificial Intelligence, Peking University} \and
Center for Data Science, Peking University \and
Yizhun Medical AI Co., Ltd  \and
Peng Cheng Laboratory \and
Pazhou Laboratory (Huangpu)
\\
\email{wangdongcis@pku.edu.cn}, 
\email{2201213301@stu.pku.edu.cn}, 
\email{zhaozw@stu.pku.edu.cn},
\email{yuhang.liu@yizhun-ai.com},
\email{chenyihong@pku.edu.cn},
\email{wanglw@cis.pku.edu.cn}
}

\maketitle

\begin{abstract}
This paper explores the point set representation for tubular structure extraction tasks. Compared with the traditional mask representation, the point set representation enjoys its flexibility and representation ability, which would not be restricted by the fixed grid as the mask. 
Inspired by this, we propose \ps, an alternative to the segmentation models for the tubular structure extraction task. \ps splits the image into scatter regions and parallelly predicts points for each scatter region. We further propose the greedy-based region-wise bipartite matching algorithm to train the network end-to-end and efficiently.
We benchmark the \ps on four public tubular datasets, and the extensive experiments on tubular structure segmentation and centerline extraction task demonstrate the effectiveness of our approach.
\emph{Code is available at \href{https://github.com/zhangzhao2022/pointscatter}{https://github.com/zhangzhao2022/pointscatter}.}

\end{abstract}

\section{Introduction}

Tubular structures broadly exist in computer vision tasks, especially medical image tasks, such as blood vessels~\cite{ma2020rose,wu2021scs}, ribs~\cite{lenga2018deep,yang2021ribseg}, and nerves~\cite{guimaraes2016fast}. Accurate extraction of these tubular structures performs a decisive role in the downstream tasks.
For instance, the diagnosis of eye-related diseases such as hypertension, diabetic retinopathy highly relies on the extraction of retinal vessels.
Deep learning-based methods usually model the extraction of tubular structures as a regular semantic segmentation task, which predicts segmentation masks as the representation of the structures.
Therefore, previous works mostly adopt the following two routines: designing novel network components to incorporate vascular or tubular priors~\cite{shin2019deep}, or proposing loss functions that promote topology preservation~\cite{shit2021cldice}. 

\begin{figure}[t]
\centering
\includegraphics[width=\textwidth]{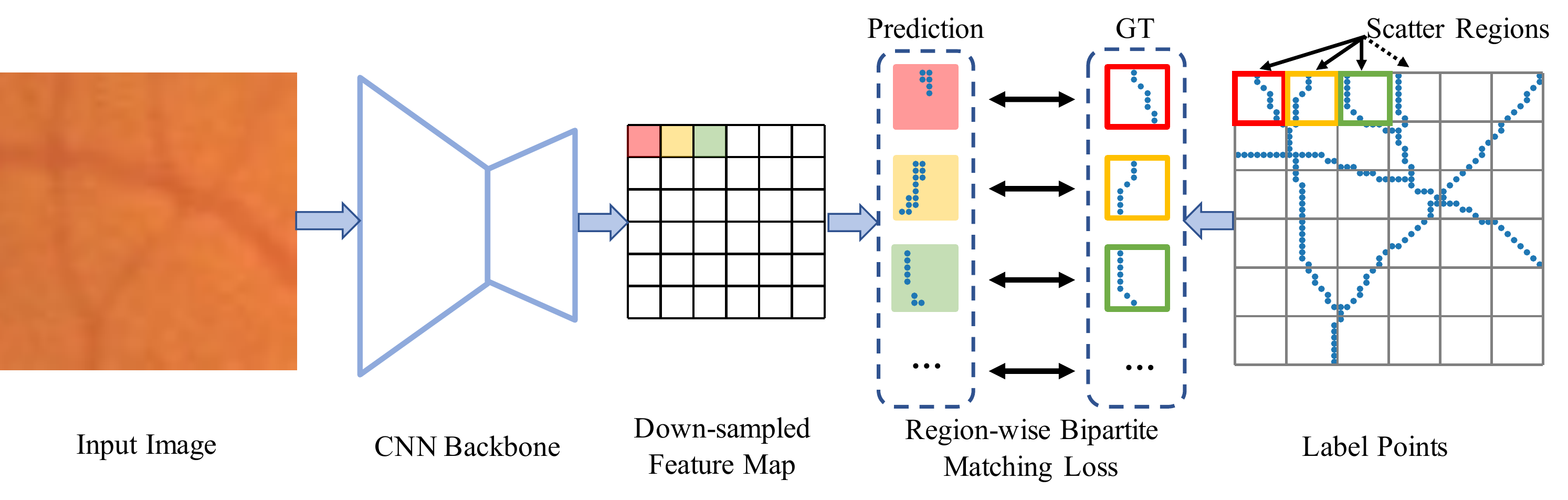}
\caption{\pointscatter adopts point set representation to perform tubular structure extraction. We exhibit a small input image with size $48 \times 48$ to show the details clearly. \ps learns to predict points for each scatter region seperately. Region-wise bipartite matching is empolyed in \pointscatter to train the network end to end.}
\label{fig:intro_framework}
\end{figure}

Semantic segmentation methods apply successively upsampling on the high-level feature maps to get the predicted segmentation masks. The wide receptive field makes it more suitable to recognize large connected areas in the image. However, the paradigm of semantic segmentation has its inherent defect, which is further amplified in the task of tubular structure extraction. It is widely known that the segmentation models struggle in extracting high-frequency information accurately, such as image contours~\cite{kirillov2020pointrend,kirillov2017instancecut,chen2017deeplab,chen2018encoder}. 
In tubular structures, the natural thinness makes almost all foreground regions contact with the structure boundaries. The special characteristics of the tubular structure increase the difficulty of capturing the fine-scale tubular details, which leads to false negatives of the small branches in the tubular structures.

We argue that the limitation lies in the representation of the prediction results. The semantic segmentation methods predict one score map to represent a segmentation result. The score map is arranged on a regular grid where each bin corresponds to a pixel in the input image. The fixed grid limits the flexibility of the representation and therefore restricts the ability of the network to learn fine-scale structures. Compared with the regular grids, point set representation is a more reasonable way for tubular structure extraction. Since the points can be placed at arbitrary real coordinates in the image, the point set representation enjoys more flexibility and expression ability to learn the detailed structures and is not restricted to a fixed grid. 

Therefore, in this paper, we propose \pointscatter to explore the feasibility of point set representation in tubular structure extraction. \pointscatter (Fig.~\ref{fig:intro_framework}) is an alternative of the mask segmentation method and can apply to regular segmentation backbones (\eg U-Net~\cite{ronneberger2015u}) with minor modifications. 
Given a downsampled feature map output by the CNN backbone network, each localization of this feature map is responsible for predicting points in the corresponding \bd{scatter region}. In this paper, we regard each patch as a scatter region as shown in Fig.~\ref{fig:intro_framework}, and each localization of the feature map corresponds to the image patch with the same relative position within the whole image. For each scatter region, our \ps predicts a fixed number of points with their objectness scores. When inference, a threshold is applied to filter out points with low scores. The aggregation of all scatter regions forms the final results.

Our \ps predicts points for all scatter regions parallelly at once, and the training process is also in an end-to-end and efficient manner. We apply the set matching approach separately for each scatter region to perform label assignment for training our \ps.
Previous works in the object detection area (\eg DETR~\cite{carion2020end}) adopt Hungarian algorithm~\cite{kuhn1955hungarian} to perform one-to-one label assignment. Following this way, a straightforward way is using the Hungarian algorithm iteratively for each scatter region. However, the iteration process is inefficient for large images with thousands of scatter regions. Consequently, we propose the region-wise bipartite matching method which is based on the greedy approach. Our method reduces the computation complexity from $O(N^3)$ to $O(N^2)$ for each scatter region and is easier to be implemented on GPU using the vectorized programming by the deep learning framework (\eg PyTorch~\cite{adam19torch}). 

The advantages of our \ps and point sets lie in their flexibility and adaptability.
1) For the segmentation methods, each pixel of the output score maps corresponds to the pixel of the input image with the same spatial location, and has to predict the objectness score for this pixel. While in our \ps, the model can adaptively decide the assignments between the predicted and GT points within each scatter region. Since there are fewer restrictions on the assignments, the model is much easier to fit the complicated fine-scale structures in the training process.
2) During the \ps training, since we use points as GT rather than the mask, the predicted points can approach the GT points along the continuous spatial dimension. The extra dimension rather than the classification score dimension will reduce the optimization difficulty and provide more paths for the optimization algorithm to find the optimal solution during model training.

Experimentally, we evaluate \ps on four typical tubular datasets. For each dataset, we compare our methods with their segmentation counterparts on three strong backbone networks. We consider two tasks for tubular structure extraction: tubular structure segmentation and centerline extraction. Extensive experimental results reveal:
\begin{enumerate}
\item On the tubular structure segmentation task, according to the volumetric scores, our \ps achieves superior performance on most of the 12 combinations of the datasets and the backbone networks.
\item On the tubular structure segmentation task, using \ps as an auxiliary task to learn the centerline, the performance of both volumetric scores and topology-based metrics of the segmentation methods will be boosted.
\item On the centerline extraction task, our \ps significantly outperforms the segmentation counterparts by a large margin.
\item The qualitative analysis shows that our \ps is better than the segmentation methods on the small branches or bifurcation points, which verifies the expression ability of our method.
\end{enumerate}

\section{Related Work}

\subsection{Tubular Structure Segmentation}
Tubular structure segmentation is a classical task due to the broad existence of tubular structures in medical images. Traditional methods~\cite{benmansour2011tubular,sironi2014multiscale,caselles1997geodesic,bauer2010segmentation,alvarez2017tracking} seek to exploit special geometric priors to improve the performance. Fethallah~\etal~\cite{benmansour2011tubular} proposes an interactive method for tubular structure extraction. Once the physicians click on a small number of points, a set of minimal paths could be obtained through the marching algorithm. Amos~\etal~\cite{sironi2014multiscale} considers centerline detection as a regression task and estimates the distance in scale space.

As for deep learning-based models, U-Net~\cite{ronneberger2015u} and FCN~\cite{long2015fully} are the classical methods for semantic segmentation, which are also appropriate for tubular structures. To further improve the performance, approaches specially designed for tubular structures have been proposed recently. These methods can be coarsely classified into two categories: incorporating tubular priors into the network architecture~\cite{wang2020deep,shin2019deep,tetteh2020deepvesselnet} and designing topology-preserving loss functions~\cite{shit2021cldice,oner2020promoting,mosinska2018beyond,hu2019topology}. Wang~\etal~\cite{wang2020deep} attempt to predict a segmentation mask and a distance map simultaneously for tubular structures. Then the mask could be refined through the shape prior reconstructed from the distance map. Shit~\etal~\cite{shit2021cldice} introduces a new similarity measure called clDice to represent the topology architecture of tubular structures. Moreover, the differentiable version soft-clDice is proposed to train arbitrary segmentation networks. Oner~\etal~\cite{oner2020promoting} proposes a connectivity-oriented loss function for training deep convolutional networks to reconstruct network-like structures. Besides these two ways, some researchers propose special approaches for their specific tasks. For instance, Li~\etal~\cite{li2021deep} leverages a deep reinforced tree-traversal agent for efficient coronary artery centerline extraction. Different from the above methods, our {\ps} is the first to utilize points as a new representation for tubular structures, which significantly improves the segmentation performance.

\subsection{Point Set Representation}

Recently, points have become a popular choice to represent objects. Contributing to its flexibility and great expression capability, point representation is applied in various fields, such as image object detection~\cite{yang2019reppoints,law2018cornernet,duan2019centernet}, instance segmentation~\cite{yang2020dense}, pose estimation~\cite{cao2017realtime,wang2020hrnet,papandreou2018personlab,zhou2021differentiable}, 3D object classification and segmentation~\cite{qi2017pointnet,qi2017pointnet++,zhao2021point}, \etc. Benefiting from the advantage of points for both localization and recognition, RepPoints~\cite{yang2019reppoints} utilizes point set as a new finer representation of objects instead of the rectangular bounding boxes. For the task of human pose estimation, detecting key points of humans is regarded as the prerequisite. Then, based on the prior knowledge of the human body, the skeletons can be obtained via the spatial connections among the detected key points. In the area of 3D object recognition, the point cloud is an important data structure. Thousands of points represented by the three coordinates (x, y, z) make up the scenes and objects. Qi~\etal~\cite{qi2017pointnet} provides a unified architecture for point cloud to achieve object classification and semantic segmentation. In this paper, we introduce the point set representation for tubular structures due to the expression ability of points to capture complex and fine-grained geometric structures.

\subsection{Set Prediction by Deep Learning}
The paradigm of set prediction has been introduced into the computer vision tasks (\eg Object Detection~\cite{carion2020end,zhu2020deformable,wang2021defcn}) firstly by DETR~\cite{carion2020end}. In DETR, a bipartite matching between ground truth and prediction is constructed based on the Hungarian algorithm~\cite{kuhn1955hungarian}, which guarantees that each target corresponds to a unique prediction. Following DETR, Wang~\etal~\cite{wang2021end} and Sun~\etal~\cite{sun2021makes} perform one-to-one label assignment for classification to enable end-to-end object detection. More recently, researchers attempt to utilize the pattern of set prediction to improve the performance of other high-level tasks~\cite{wang2021max,cheng2021per,wang2021end,zou2021end,chen2021reformulating}. Cheng~\etal~\cite{cheng2021per} reformulates semantic segmentation as a mask set prediction problem and shows excellent empirical results. Wang~\etal~\cite{wang2021end} predicts instance sequences directly via instance sequence set matching for video instance segmentation. In~\cite{zou2021end,chen2021reformulating}, the HOI instances can make up the triplet instance sets for both ground truth and prediction, which provides a simple and effective manner for Human Object Interaction (HOI) detection. 
Moreover, in the task of instance-aware human part parsing, \cite{zhou2021differentiable} designs a specific differentiable matching method to generate the matching results for predicted limbs with different categories. 
In this paper, to train our \ps, we divide the image by predefined scatter regions and perform set predictions between predicted points and GT points on each of the regions in parallel.

\section{Methodology}

\begin{figure}[t]
\centering
\includegraphics[width=\textwidth]{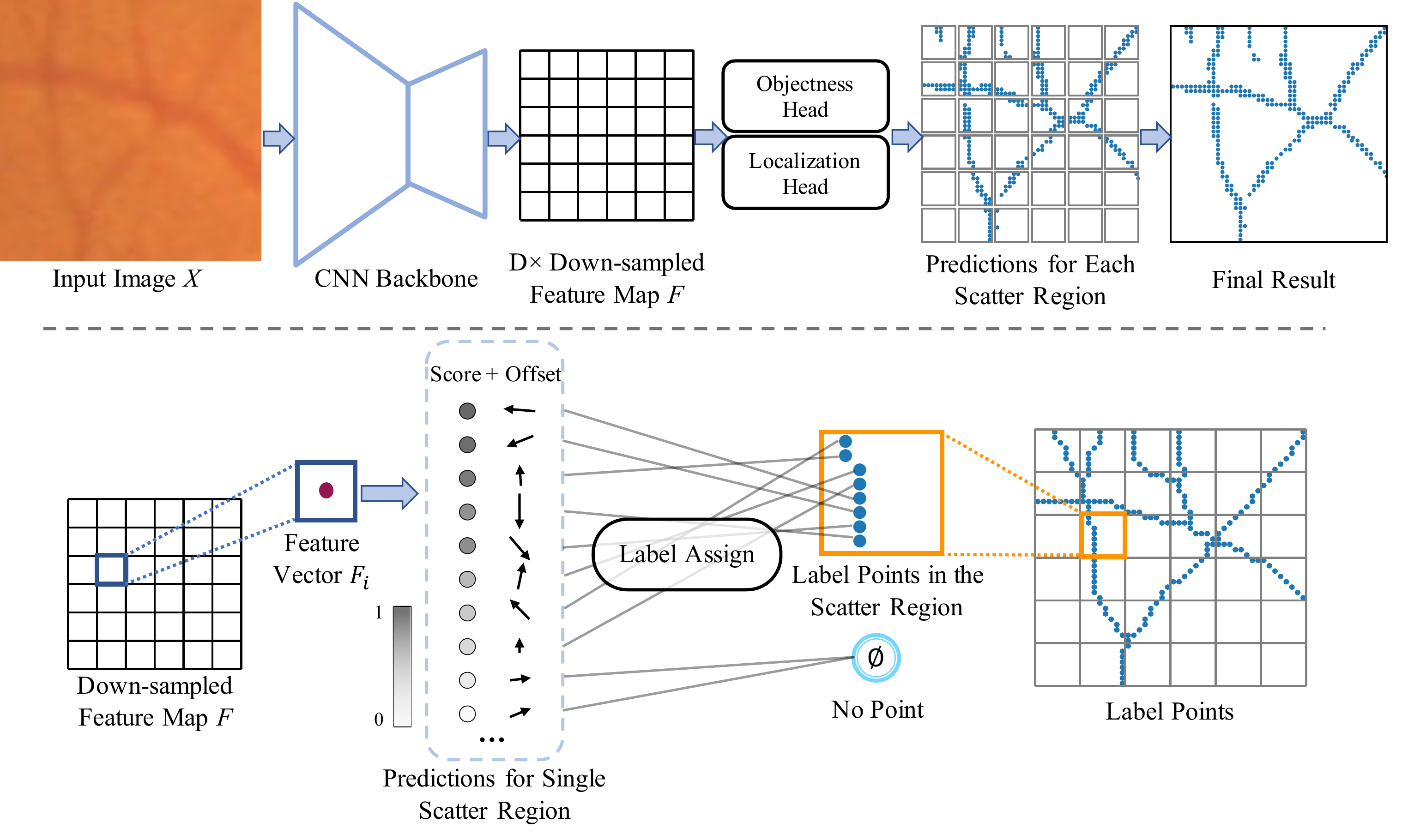}
\caption{\bd{The pipeline of \pointscatter.} The top part illustrates the pipeline of point set prediction of \pointscatter. It predicts points for each scatter region separately and gathers them to form the final result. The bottom part exhibits the approach of label assignment for each scatter region. We obtain point-to-point assignments to supervise the network training precisely. The predicted points without match will be allocated a ``no point'' class.}
\label{fig:main}
\end{figure}

The \ps receives 2D images and produces point sets to represent the tubular structure. The training process of the set prediction task is end-to-end and efficient contributed by the region-wise bipartite matching method. We will first introduce the architecture of \ps in Section~\ref{sec:arch}. Then Section~\ref{sec:train} elaborates on the training process. An overview of our proposed \ps is shown in Fig~\ref{fig:main}.

\subsection{\pointscatter Architecture}
\label{sec:arch}

Our \pointscatter formulates the tubular structure extraction task as a point set prediction task. The pipeline of the inference process of \pointscatter is illustrated in the top part of Fig.~\ref{fig:main}. Given an input image $X$ with shape $\mathbb{H} \times \mathbb{W}$, it is firstly fed into the CNN backbone network, and we obtain the corresponding down-sampled feature map $F \in \mathbb{R}^{C \times H \times W}$, where $C$ is the channel size, $H$ and $W$ indicate the shape of the feature map. Let $D$ denote the downsampling rate of the CNN backbone, we have
\begin{equation}
H = \mathbb{H} / D,\quad W = \mathbb{W} / D.
\end{equation}
Note that we assume that $\mathbb{H}$ and $\mathbb{W}$ are divisible by $D$, which is the same situation as semantic segmentation. 

Next, we introduce the concept of \bd{scatter region}. In \ps, each spatial localization $F_i$ in $F$ is responsible for predicting the corresponding points that situate in a predefined region of the input image. $i$ denotes the spatial index in $H \times W$. We call this predefined specific area \bd{scatter region}. Note that the scatter region could be of arbitrary shape. In this paper, considering the natural grid shape of the feature map $F$, we define the scatter region as the $D \times D$ patch which has the same relative position in the input image as $F_i$ in $F$. The top part of Fig.~\ref{fig:main} provides an intuitive illustration.

We employ two head networks to perform point prediction for each scatter region. The objectness head and the localization head are responsible for producing point scores and offsets, respectively. The points of all scatter regions jointly constitute the final output.

The points prediction mechanism from a high-level feature map $F$ makes our \ps different from the mask segmentation methods (\eg U-Net~\cite{ronneberger2015u}). Instead of generating a grid of mask with the same shape as the original image, our \ps utilizes flexible points to describe the tubular structures. The ampliative representation ability enhances the power of the network to learn the complicated fine-scale structures. We will then introduce the details of the head and backbone networks in the following.

\begin{figure}[t]

\sbox\twosubbox{%
  \resizebox{\dimexpr0.98\textwidth-1em}{!}{%
    \includegraphics[height=5cm]{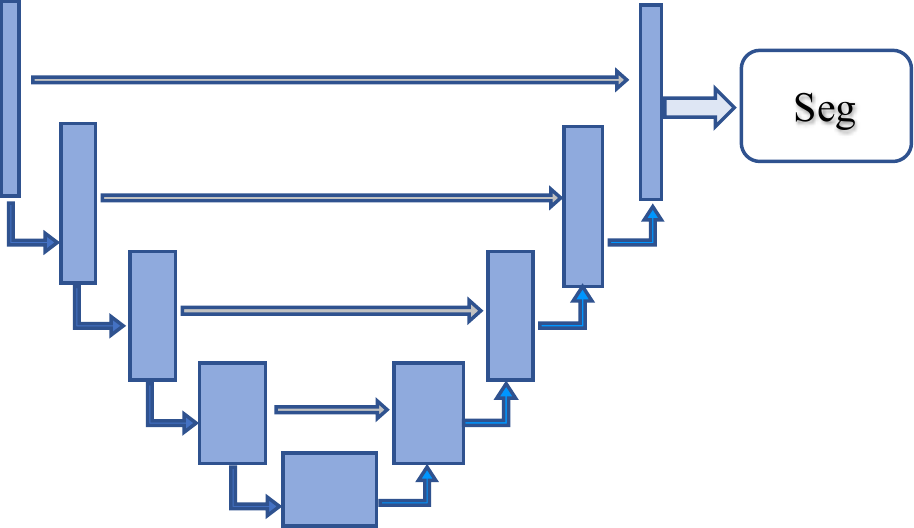}%
    \includegraphics[height=5cm]{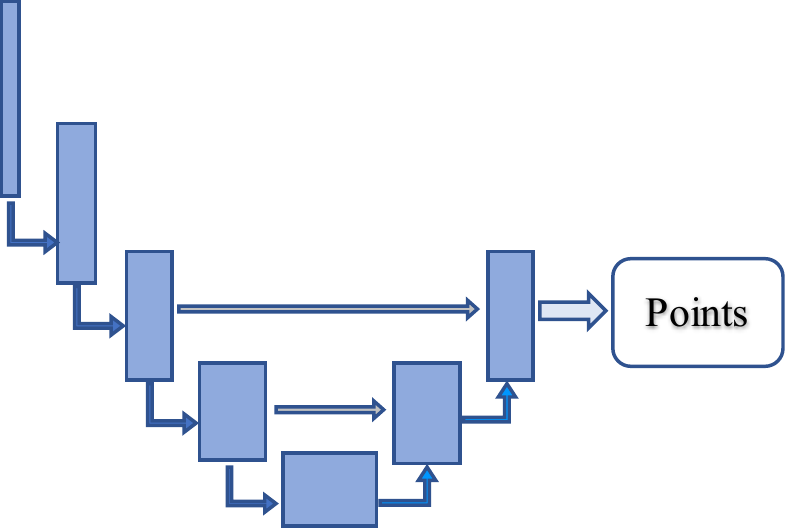}%
    \includegraphics[height=5cm]{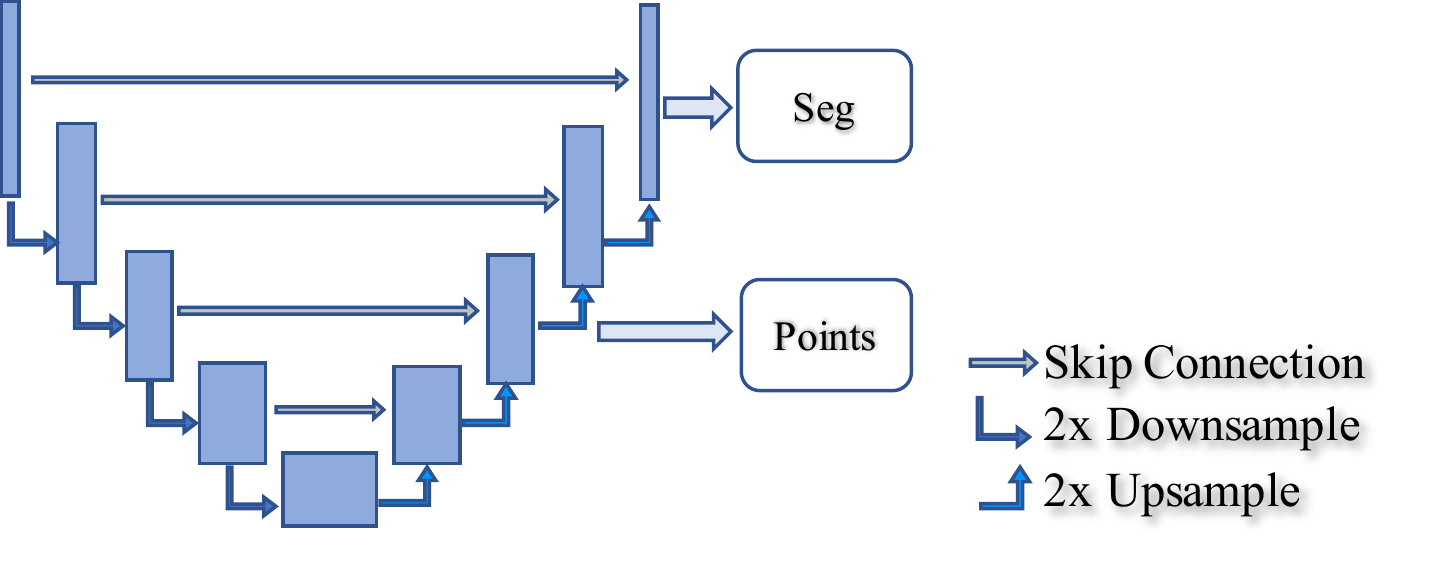}%
  }%
}
\setlength{\twosubht}{\ht\twosubbox}

\centering
\subcaptionbox{Segmentation\label{fig:unet-seg}}{%
  \includegraphics[height=\twosubht]{figs/unet1}%
}\quad
\subcaptionbox{\ps\label{fig:unet-ps}}{%
  \includegraphics[height=\twosubht]{figs/unet2}%
}
\subcaptionbox{Segmentation + \ps \label{fig:unet-seg-ps}}{%
  \includegraphics[height=\twosubht]{figs/unet3}%
}
\caption{Illustrations of applying segmentation method or our \ps on the U-Net backbone. We show an abstractive version of U-Net for ease of presentation. We set the downsample scale $D=4$.}
\label{fig:unet}
\end{figure}

\subsubsection{Head Networks. }
The head networks are responsible for predicting points for each image scatter region. They are composed of the Objectness Head (ObjHead) and the Localization Head (LocHead).
For each scatter region, \ps generates $N$ point candidates with their objectness scores and their localization, where $N$ should be set greater than the maximum number of label points within a scatter region.

Formally, given $F_i$ from the downsampled feature map $F$, where $i$ indicates the spatial localization in $H \times W$, we denote the center localization of the corresponding image scatter region in the input image as $c_i = (X_i^c, Y_i^c)$. Then we predict the objectness score for $N$ points and regress their coordination offsets relative to $c_i$ by ObjHead and LocHead, respectively:
\begin{equation}
\text{score}_i = \text{Sigmoid}(\text{ObjHead}(F_i)) \in \mathbb{R}^N, \quad \text{offset}_i= \text{LocHead}(F_i) \in \mathbb{R}^{N \times 2}.
\end{equation}
Note that we apply Sigmoid operation after ObjHead to normalize the objectness score to the scale of $[0, 1]$. To acquire the coordinate of the points, we can simply apply the regressed offsets to the center point $c_i$, \ie $p_i^j=(X_i^c + \text{offset}_i^{j,1}, Y_i^c + \text{offset}_i^{j,2})$, where $p_i^j$ is the $j_\text{th}$ point generated from $F_i$. During inference, the points with scores lower than a threshold $T$ will be eliminated. Sliding these two heads on the whole feature map $F$ and merging all predicted points, we can obtain the final results.

In practice, we instantiate ObjHead and LocHead both as a single linear layer, which can be implemented by the convolutional layer with kernel size $1 \times 1$. Although the transformer-like architecture~\cite{vaswani2017attention} is proven to promote the interaction between object items in prior works~\cite{carion2020end,zhu2020deformable}, to maintain simplicity and focus on the point representation itself, we adopt the fully convolutional architecture in our \ps.

\subsubsection{Backbone Networks. }
As introduced above, in \ps, the only requirement on the backbone network is that it should produce a $D \times$ downsampled feature map relative to the input image. The universality of \ps makes it compatible with almost all common backbone networks in semantic segmentation~\cite{long2015fully,ronneberger2015u}. In this section, we take as an example the most famous model for medical image segmentation U-Net~\cite{ronneberger2015u} to show how we apply \ps to a regular segmentation network.

We illustrate the utilization of the backbone network (\ie U-Net) in Fig.~\ref{fig:unet}. Traditional segmentation methods use the feature map with the same shape as the input image to generate the corresponding segmentation mask (Fig.~\ref{fig:unet-seg}). Our \ps (Fig.~\ref{fig:unet-ps}) passes the $D \times$ downsampled feature map to the head networks, while the successive upsampled feature maps are removed from the computational graph. In Fig.~\ref{fig:unet-seg-ps}, we show that we can simultaneously apply segmentation and \ps to the backbone network, which can be regarded as the multitask learning manner. We find that multitask learning will boost the performance of mask segmentation in the following experiments section.

\subsection{Training \pointscatter}
\label{sec:train}

The training of \ps also complies with the paradigm of scatter regions. Therefore, for each scatter region, we should assign the class label and the offset label for each predicted point. To achieve this goal, following prior works~\cite{carion2020end,zhu2020deformable}, we first define the cost function between predicted and ground-truth points, and then perform bipartite matching to produce one-to-one label assignment with a low global cost (bottom part of Fig.~\ref{fig:main}).

We first discuss the cost function. The matching cost should take into account both the objectness scores and the distance of the predicted and ground-truth points. Specifically, for each scatter region, we have a set of $K$ ground-truth points $G=\{g_i\}_{i=1}^K$ and $N$ predicted points $P=\{p_i\}_{i=1}^N$. Note that we omit the index of scatter region in this subsection for simplicity.
For each predicted point $p_i$, its objectness score is denoted as $s_i$. Since the common assumption is $K \le N$, we consider $G$ also a set of size $N$, where the rest part is complemented by $\varnothing$ (no point). Therefore, for a permutation of $N$ elements $\sigma \in \mathfrak{S}_N$, we define the cost for each point assignment as
\begin{equation}
{\cal L}_{\rm match} (g_i,p_{\sigma(i)}) = [ L_1 (g_i,p_{\sigma(i)})]^\eta \cdot \lvert s_{\sigma(i)} - \mathds{1}(i \le K) \rvert ^{1-\eta} ,
\label{equ:cost}
\end{equation}
where the first term in the equation describes the matching quality of point localization, and the last item indicates the classification error. $L_1$ is the manhattan distance in this equation, and $\eta$ is a hyper-parameter determined by cross-validation. Note that we use multiplication instead of addition across the two cost terms, since the effectiveness of multiplication has been proven in \cite{defcn_2021_CVPR}. 

\subsubsection{Label Assignment with Region-wise Bipartite Matching.} The second step is to get the optimal permutation $\sigma$. Previous works such as DETR~\cite{carion2020end} adopt the Hungarian algorithm~\cite{kuhn1955hungarian} to perform set matching. Following this way, a direct generalization to our problem is to compute the bipartite matching iteratively for each scatter region. However, due to the large number of scatter regions in the images, it is inefficient to execute the iteration.

To tackle this problem, we propose a greedy-based bipartite matching method. We present the matching for each image scatter region in Algorithm~\ref{alg:greedy}. The greedy bipartite matching iterates the ground-truth points and finds the predicted point with the minimum cost from the left predicted points. The greedy method reduces the computational complexity of the Hungarian algorithm from $O(N^3)$ to $O(N^2)$, and is easy to be implemented on GPU using the deep learning framework (\ie PyTorch) for batched computation. We provide an efficient implementation of the greedy bipartite matching in the supplementary materials.

The greedy method could not generate the optimal matching results. However, the network predictions become gradually closer to the ground-truth points during training. The optimization of the network improves the quality of predicted points, which makes the matching problem easier, hence the weak greedy method is capable of allocating the point targets.

\begin{algorithm}[h]
  \caption{Greedy Bipartite Matching}
  \label{alg:greedy}
\begin{algorithmic}[1]
  \STATE {\bfseries Input:} $G=\{g_1, g_2, ..., g_K\}$, $P=\{p_1, p_2, ..., p_N\}$, $C \in \mathbb{R}^{K \times N}$
  \STATE $G$ is the list of ground truth points, $P$ is the list of predicted points, $C$ is the cost matrix, $C_{i \cdot}$ is the i-th row of $C$, $C_{\cdot j}$ is the j-th column of $C$
  \STATE {\bfseries Output:} $S=\{\sigma(1), \sigma(2), ..., \sigma(K)\}$, $\sigma(i)$ represents that the predicted point $p_{\sigma(i)}$ is assigned to the ground truth point $g_i$; the rest of predicted points $\{p_{\sigma(K+1)},...,p_{\sigma(N)}\}$ are assigned to ``no point''
  \STATE {\bfseries begin}
  \STATE \ \ \ \ {\bfseries for} $i=1$ {\bfseries to} $K$
  \STATE \ \ \ \ \ \ \ \ $n \leftarrow$ argmin $C_{i\cdot}$
  \STATE  \ \ \ \ \ \ \ \  $\sigma(i) \leftarrow n$
  \STATE \ \ \ \ \ \ \ \ $C_{\cdot n} \leftarrow$ inf
  \STATE \ \ \ \ {\bfseries end}
  \STATE \ \ \ \ {\bfseries return} $S$
  \STATE {\bfseries end}
\end{algorithmic}
\end{algorithm}

\subsubsection{Loss Functions.} To train \ps, the loss function is composed of objectness loss and regression loss:
\begin{equation}
\mathcal{L}_{\text{total}} = \mathcal{L}_{\text{obj}} + \lambda \mathcal{L}_{\text{reg}},
\label{equ:loss}
\end{equation}
where $\mathcal{L}_{\text{obj}}$ is instantiated as Focal Loss~\cite{lin2017focal} to deal with the unbalanced distribution of objectness targets and we use $\text{L}_1$ loss for $\mathcal{L}_{\text{reg}}$. Note that the regression loss is only applied to the positive points and the points matched ``no point'' will be eliminated.
Practically, we normalize the total loss by dividing the number of ground-truth points to keep the optimization process stable.

It is worth mentioning that almost all current datasets for tubular structure extraction provide mask annotation, therefore we should convert the mask annotation to points to supervise the training of \ps. We accomplish this goal by replacing each pixel with one point located in the center of this pixel.
Concretely, a mask is represented as a matrix with binary values $\mathbf{Y}^{\mathbb{H} \times \mathbb{W}} \in \{ 0, 1 \}$, and we convert it to the point sets $\{ (i, j) | Y_{i, j}=1, i \in [1, \mathbb{H}], j \in [1, \mathbb{W}] \}$.

\section{Experiments}

\subsection{Experimental setup}

\subsubsection{Datasets.}
We evaluate our \ps on four public tubular datasets, including two medical datasets and two satellite datasets. DRIVE~\cite{staal2004ridge} and STARE~\cite{hoover2000locating} are two retinal datasets that are commonly adopted in the medical image segmentation problem to evaluate the performance of vessel segmentation. The Massachusetts Roads (MassRoad) dataset~\cite{mnih2013machine} and DeepGlobe~\cite{demir2018deepglobe} are labeled with the pixel-level annotation for road segmentation. We use the official data split for DRIVE and STARE in MMSegmentation~\cite{mmseg2020} and follow the data split method in previous works~\cite{shit2021cldice,singh2018self} for the other two datasets. We report the performance on the test set.

\subsubsection{Tasks and Metrics.}
In this paper, we focus on two different tasks relevant to the understanding of tubular structures: tubular structure segmentation and centerline extraction. The above four datasets are used for the image segmentation task in previous works, and the most popular dataset for centerline extraction is the MICCAI 2008 Coronary Artery Tracking (CAT08) dataset~\cite{schaap2009standardized}. But unfortunately, the CAT08 dataset and the evaluation server are not publicly available now. Therefore, we generate the centerline labels using the skeleton extraction method in \cite{shit2021cldice} to fulfill the evaluation of centerline extraction. The labelled centerline is a set of connected pixels with a line-like structure where the width is 1 pixel. It is a more challenging task to extract the centerlines accurately by deep models. We then introduce the metrics we utilize for these two tasks.

For the tubular structure segmentation task, we consider two types of metrics: volumetric and topology-based. The volumetric scores include Dice coefficient, Accuracy, AUC, and the recently proposed clDice~\cite{shit2021cldice}. We also calculate the topology-based scores including the mean of absolute Betti Errors for the Betti Numbers $\beta_0$ and $\beta_1$ and the mean absolute error of Euler characteristic. We follow \cite{shit2021cldice} to compute the topology-based scores.

For the centerline extraction task, we report the Dice coefficient, Accuracy, AUC, Precision, and Recall as the volumetric scores. To increase the robustness of the evaluation, we apply a three-pixel tolerance region around the centerline annotation following \cite{guimaraes2016fast}. We adopt the same topology-based metrics as the tubular structure segmentation task.

The above metrics are designed for mask prediction, while our \ps generates points to describe the foreground structures. Therefore, we should convert the points to the segmentation mask in order to accommodate the evaluation protocol. Specifically, an image can be regarded as a grid with the size of each bin $1 \times 1$, and each point is expected to be located in one bin. We first initialize an empty score map with the same size as the input image. For each bin in the output score map, we directly set the objectness score of the point located in this bin as its score. The bins without any point will be endowed with zero scores. To get the segmentation mask, we can threshold the score map by 0.5.

\subsubsection{Implementation Details.}
As discussed in Section~\ref{sec:arch}, our \ps is compatible with various segmentation backbones with an encoder-decoder shape, the adjustable parameters are the downsample rate $D$ and the number of points in each scatter region $N$. Experimentally, we set $D=4$ and $N=16$ by default. The threshold of objectness score during inference is $T=0.1$.
During training, we set $\eta=0.8$ in Equation.~\ref{equ:cost} to balance the localization cost and classification cost. 
Additional implementation details are depicted in the supplementary materials. We implement our model based on PyTorch and MMSegmentation~\cite{mmseg2020}.

\begin{table}[t]
\centering
\caption{Main results on tubular structure segmentation task. The gray lines use our \ps. We mark the best performance by bold numbers.} 
\label{tab:seg}
\renewcommand\arraystretch{1.2}
\scalebox{0.49}{
\begin{tabular}{c | l | l | c c c c | c c c}
\toprule
\mrtwo{Dataset}	&\mrtwo{Backbone}	&\mrtwo{Method}	& \multicolumn{4}{c|}{Volumetric Scores (\%) $\uparrow$}& \multicolumn{3}{c}{Topological Error $\downarrow$} \\
\cline{4-10}
&&&AUC	&Dice	&clDice	&ACC	&$\beta_0$	&$\beta_1$	&$\chi_{error}$	\\
\midrule										
	&	&softDice	&97.05	&81.09	&80.69	&95.28	&1.504	&1.129	&1.806	\\
	&	&clDice	&96.84	&81.15	&81.55	&95.21	&1.072	&0.993	&1.354	\\
\rowcolor{mygray}\cellcolor{white}{}&\cellcolor{white}{UNet}
		&\ps	&\bd{97.69}	&\bd{81.63}	&\bd{82.89}	&95.23	&1.317	&1.250	&1.628	\\
\rowcolor{mygray}\cellcolor{white}{}&\cellcolor{white}{}
		&softDice+\psaux	&97.27	&81.59	&81.43	&\bd{95.37}	&1.004	&0.980	&1.269	\\
\rowcolor{mygray}\cellcolor{white}{}&\cellcolor{white}{}
		&clDice+\psaux	&96.97	&81.51	&82.54	&95.24	&\bd{0.873}	&\bd{0.944}	&\bd{1.131}	\\
\cline{2-10}										
	&	&softDice	&96.42	&80.96	&80.55	&95.24	&1.698	&1.106	&1.978	\\
	&	&clDice	&96.77	&81.10	&81.48	&95.17	&1.105	&0.965	&1.359	\\
\rowcolor{mygray}\cellcolor{white}{DRIVE}&\cellcolor{white}{UNet++}
		&\ps	&\bd{97.45}	&\bd{81.38}	&\bd{82.34}	&95.17	&1.290	&1.225	&1.600	\\
\rowcolor{mygray}\cellcolor{white}{}&\cellcolor{white}{}
		&softDice+\psaux	&96.45	&81.31	&81.03	&\bd{95.29}	&0.936	&0.956	&\bd{1.184}	\\
\rowcolor{mygray}\cellcolor{white}{}&\cellcolor{white}{}
		&clDice+\psaux	&96.51	&81.28	&81.62	&95.22	&\bd{0.924}	&\bd{0.937}	&1.189	\\
\cline{2-10}										
	&	&softDice	&97.78	&82.11	&82.28	&95.49	&1.284	&1.067	&1.562	\\
	&	&clDice	&97.09	&81.43	&82.48	&95.21	&1.005	&\bd{1.006}	&1.272	\\
\rowcolor{mygray}\cellcolor{white}{}&\cellcolor{white}{ResUNet}
		&\ps	&97.87	&81.85	&82.75	&95.34	&1.547	&1.273	&1.834	\\
\rowcolor{mygray}\cellcolor{white}{}&\cellcolor{white}{}
		&softDice+\psaux	&\bd{97.97}	&\bd{82.45}	&82.64	&\bd{95.59}	&1.372	&1.023	&1.628	\\
\rowcolor{mygray}\cellcolor{white}{}&\cellcolor{white}{}
		&clDice+\psaux	&97.36	&82.02	&\bd{84.62}	&95.31	&\bd{0.883}	&1.019	&\bd{1.142}	\\
\midrule										
	&	&softDice	&94.86	&82.27	&84.87	&97.45	&1.093	&0.667	&1.260	\\
	&	&clDice	&96.82	&82.29	&85.22	&97.44	&0.790	&0.665	&0.943	\\
\rowcolor{mygray}\cellcolor{white}{}&\cellcolor{white}{UNet}
		&\ps	&\bd{97.86}	&82.73	&85.83	&97.45	&0.818	&0.774	&0.978	\\
\rowcolor{mygray}\cellcolor{white}{}&\cellcolor{white}{}
		&softDice+\psaux	&96.42	&82.78	&85.44	&97.51	&0.727	&0.625	&0.887	\\
\rowcolor{mygray}\cellcolor{white}{}&\cellcolor{white}{}
		&clDice+\psaux	&97.32	&\bd{83.11}	&\bd{86.45}	&\bd{97.54}	&\bd{0.631}	&\bd{0.614}	&\bd{0.778}	\\
\cline{2-10}										
	&	&softDice	&95.05	&82.22	&84.60	&97.45	&1.005	&0.667	&1.163	\\
	&	&clDice	&96.48	&82.62	&85.72	&97.49	&0.801	&0.648	&0.968	\\
\rowcolor{mygray}\cellcolor{white}{STARE}&\cellcolor{white}{UNet++}
		&\ps	&\bd{97.85}	&82.80	&85.98	&97.43	&0.844	&0.745	&0.997	\\
\rowcolor{mygray}\cellcolor{white}{}&\cellcolor{white}{}
		&softDice+\psaux	&95.59	&82.85	&85.54	&\bd{97.53}	&0.658	&0.649	&0.801	\\
\rowcolor{mygray}\cellcolor{white}{}&\cellcolor{white}{}
		&clDice+\psaux	&96.36	&\bd{82.96}	&\bd{86.11}	&\bd{97.53}	&\bd{0.650}	&\bd{0.617}	&\bd{0.800}	\\
\cline{2-10}										
	&	&softDice	&96.27	&81.65	&84.11	&97.38	&0.913	&0.695	&1.051	\\
	&	&clDice	&96.65	&82.51	&85.33	&97.47	&0.731	&0.650	&0.884	\\
\rowcolor{mygray}\cellcolor{white}{}&\cellcolor{white}{ResUNet}
		&\ps	&\bd{97.77}	&82.40	&85.00	&97.38	&0.949	&0.730	&1.093	\\
\rowcolor{mygray}\cellcolor{white}{}&\cellcolor{white}{}
		&softDice+\psaux	&96.59	&81.80	&83.55	&97.41	&0.796	&0.670	&0.944	\\
\rowcolor{mygray}\cellcolor{white}{}&\cellcolor{white}{}
		&clDice+\psaux	&96.04	&\bd{82.68}	&\bd{85.60}	&\bd{97.48}	&\bd{0.601}	&\bd{0.636}	&\bd{0.748}	\\
\bottomrule										
\end{tabular}
}
\scalebox{0.49}{
\begin{tabular}{c | l | l | c c c c | c c c}
\toprule
\mrtwo{Dataset}	&\mrtwo{Backbone}	&\mrtwo{Method}	& \multicolumn{4}{c|}{Volumetric Scores (\%) $\uparrow$}& \multicolumn{3}{c}{Topological Error $\downarrow$} \\
\cline{4-10}
&&&AUC	&Dice	&clDice	&ACC	&$\beta_0$	&$\beta_1$	&$\chi_{error}$	\\
\midrule										
	&	&softDice	&97.02	&76.96	&86.33	&96.86	&0.686	&1.361	&1.356	\\
	&	&clDice	&95.76	&76.11	&85.68	&96.68	&0.679	&1.380	&1.334	\\
\rowcolor{mygray}\cellcolor{white}{}&\cellcolor{white}{UNet}
		&\ps	&\bd{97.65}	&77.57	&86.42	&96.87	&0.944	&1.353	&1.616	\\
\rowcolor{mygray}\cellcolor{white}{}&\cellcolor{white}{}
		&softDice+\psaux	&97.59	&\bd{78.14}	&\bd{87.38}	&\bd{96.98}	&0.526	&\bd{1.257}	&1.190	\\
\rowcolor{mygray}\cellcolor{white}{}&\cellcolor{white}{}
		&clDice+\psaux	&96.60	&77.68	&87.34	&96.88	&\bd{0.498}	&1.316	&\bd{1.187}	\\
\cline{2-10}										
	&	&softDice	&97.10	&76.88	&86.08	&96.82	&0.690	&1.373	&1.351	\\
	&	&clDice	&95.80	&76.39	&86.15	&96.72	&0.685	&1.455	&1.373	\\
\rowcolor{mygray}\cellcolor{white}{MassRoads}&\cellcolor{white}{UNet++}
		&\ps	&\bd{97.62}	&77.65	&86.40	&96.90	&0.836	&1.315	&1.503	\\
\rowcolor{mygray}\cellcolor{white}{}&\cellcolor{white}{}
		&softDice+\psaux	&97.60	&\bd{78.10}	&87.24	&\bd{96.99}	&0.559	&\bd{1.306}	&1.252	\\
\rowcolor{mygray}\cellcolor{white}{}&\cellcolor{white}{}
		&clDice+\psaux	&96.41	&77.81	&\bd{87.34}	&96.91	&\bd{0.498}	&1.316	&\bd{1.181}	\\
\cline{2-10}										
	&	&softDice	&96.93	&76.04	&85.57	&96.73	&0.992	&1.478	&1.658	\\
	&	&clDice	&96.12	&75.97	&85.69	&96.68	&0.887	&1.521	&1.571	\\
\rowcolor{mygray}\cellcolor{white}{}&\cellcolor{white}{ResUNet}
		&\ps	&\bd{97.40}	&76.34	&85.07	&96.74	&1.448	&1.423	&2.039	\\
\rowcolor{mygray}\cellcolor{white}{}&\cellcolor{white}{}
		&softDice+\psaux	&\bd{97.40}	&\bd{77.08}	&86.31	&\bd{96.85}	&\bd{0.745}	&\bd{1.416}	&\bd{1.435}	\\
\rowcolor{mygray}\cellcolor{white}{}&\cellcolor{white}{}
		&clDice+\psaux	&96.69	&76.67	&\bd{86.36}	&96.76	&0.803	&1.456	&1.472	\\
\midrule										
	&	&softDice	&97.65	&74.71	&80.08	&97.89	&1.154	&0.605	&1.166	\\
	&	&clDice	&96.03	&74.96	&81.16	&97.90	&0.691	&0.556	&0.751	\\
\rowcolor{mygray}\cellcolor{white}{}&\cellcolor{white}{UNet}
		&\ps	&\bd{98.64}	&78.07	&82.38	&98.12	&0.855	&0.541	&0.907	\\
\rowcolor{mygray}\cellcolor{white}{}&\cellcolor{white}{}
		&softDice+\psaux	&98.27	&\bd{78.09}	&\bd{83.96}	&\bd{98.15}	&0.492	&\bd{0.449}	&0.530	\\
\rowcolor{mygray}\cellcolor{white}{}&\cellcolor{white}{}
		&clDice+\psaux	&96.87	&77.20	&83.31	&98.07	&\bd{0.435}	&0.472	&\bd{0.485}	\\
\cline{2-10}										
	&	&softDice	&97.51	&75.64	&81.58	&97.95	&0.704	&0.549	&0.763	\\
	&	&clDice	&97.03	&75.63	&82.03	&97.94	&0.590	&0.646	&0.706	\\
\rowcolor{mygray}\cellcolor{white}{DeepGlobe}&\cellcolor{white}{LinkNet34}
		&\ps	&\bd{98.59}	&\bd{79.21}	&84.04	&98.20	&0.710	&0.543	&0.802	\\
\rowcolor{mygray}\cellcolor{white}{}&\cellcolor{white}{}
		&softDice+\psaux	&98.00	&78.88	&\bd{85.43}	&\bd{98.23}	&0.491	&\bd{0.446}	&0.549	\\
\rowcolor{mygray}\cellcolor{white}{}&\cellcolor{white}{}
		&clDice+\psaux	&97.55	&78.58	&85.08	&98.18	&\bd{0.451}	&0.448	&\bd{0.516}	\\
\cline{2-10}										
	&	&softDice	&97.45	&75.60	&81.59	&97.95	&0.604	&0.524	&0.655	\\
	&	&clDice	&97.07	&75.23	&82.18	&97.92	&0.938	&0.562	&1.009	\\
\rowcolor{mygray}\cellcolor{white}{}&\cellcolor{white}{DinkNet34}
		&\ps	&\bd{98.66}	&\bd{79.39}	&84.36	&98.20	&0.749	&0.558	&0.833	\\
\rowcolor{mygray}\cellcolor{white}{}&\cellcolor{white}{}
		&softDice+\psaux	&98.30	&78.95	&\bd{85.33}	&\bd{98.21}	&0.513	&\bd{0.440}	&\bd{0.571}	\\
\rowcolor{mygray}\cellcolor{white}{}&\cellcolor{white}{}
		&clDice+\psaux	&97.78	&78.29	&85.01	&98.16	&\bd{0.511}	&0.549	&0.613	\\
\bottomrule										
\end{tabular}
}
\end{table}

\begin{table}
\centering
\caption{Main results on centerline extraction task.} 
\label{tab:cline}
\renewcommand\arraystretch{1.2}
\scalebox{0.465}{
\begin{tabular}{c | l | l | c c c c c | c c c}
\toprule
\mrtwo{Dataset}	&\mrtwo{Backbone}	&\mrtwo{Method}	& \multicolumn{5}{c|}{Volumetric Scores (\%) $\uparrow$}& \multicolumn{3}{c}{Topological Error $\downarrow$}\\
\cline{4-11}
&&	&AUC	&Dice	&Prec	&Recall	&ACC	&$\beta_0$	&$\beta_1$	&$\chi_{error}$	\\
\midrule											
	&	&softDice	&89.53	&73.41	&90.97	&61.52	&97.63	&3.177	&1.843	&3.555	\\
\rowcolor{mygray}\cellcolor{white}{}&\cellcolor{white}{UNet}
		&\ps	&\bd{94.46}	&\bd{81.92}	&\bd{92.52}	&\bd{73.51}	&\bd{98.05}	&5.203	&2.612	&5.509	\\
\rowcolor{mygray}\cellcolor{white}{}&\cellcolor{white}{}
		&softDice+{\texttt{PS}\xspace}	&93.13	&75.92	&91.08	&65.08	&97.76	&\bd{2.677}	&\bd{1.753}	&\bd{3.051}	\\
\cline{2-11}											
	&	&softDice	&83.83	&72.06	&90.48	&59.87	&97.54	&5.651	&2.371	&6.006	\\
\rowcolor{mygray}\cellcolor{white}{DRIVE}&\cellcolor{white}{UNet++}
		&\ps	&\bd{93.29}	&\bd{82.23}	&91.14	&\bd{74.91}	&\bd{98.04}	&\bd{1.959}	&\bd{1.657}	&\bd{2.282}	\\
\rowcolor{mygray}\cellcolor{white}{}&\cellcolor{white}{}
		&softDice+{\texttt{PS}\xspace}	&87.88	&72.93	&\bd{91.29}	&60.72	&97.61	&6.360	&2.600	&6.671	\\
\cline{2-11}											
	&	&softDice	&90.83	&74.86	&91.36	&63.40	&97.70	&2.774	&1.723	&3.147	\\
\rowcolor{mygray}\cellcolor{white}{}&\cellcolor{white}{ResUNet}
		&\ps	&94.79	&\bd{83.91}	&91.73	&\bd{77.31}	&\bd{98.18}	&3.169	&1.983	&3.495	\\
\rowcolor{mygray}\cellcolor{white}{}&\cellcolor{white}{}
		&softDice+{\texttt{PS}\xspace}	&\bd{95.40}	&81.52	&\bd{92.95}	&72.60	&98.12	&\bd{2.479}	&\bd{1.673}	&\bd{2.853}	\\
\midrule											
	&	&softDice	&86.99	&72.14	&\bd{93.01}	&58.91	&98.91	&3.053	&1.562	&3.253	\\
\rowcolor{mygray}\cellcolor{white}{}&\cellcolor{white}{UNet}
		&\ps	&\bd{94.52}	&\bd{81.77}	&92.09	&\bd{73.52}	&\bd{99.10}	&2.158	&1.424	&2.303	\\
\rowcolor{mygray}\cellcolor{white}{}&\cellcolor{white}{}
		&softDice+{\texttt{PS}\xspace}	&88.34	&75.10	&92.51	&63.20	&98.98	&\bd{1.870}	&\bd{1.219}	&\bd{2.061}	\\
\cline{2-11}											
	&	&softDice	&84.94	&73.08	&92.87	&60.24	&98.93	&3.388	&1.556	&3.588	\\
\rowcolor{mygray}\cellcolor{white}{STARE}&\cellcolor{white}{UNet++}
		&\ps	&\bd{93.07}	&\bd{80.03}	&92.74	&\bd{70.38}	&\bd{99.07}	&2.582	&1.613	&2.743	\\
\rowcolor{mygray}\cellcolor{white}{}&\cellcolor{white}{}
		&softDice+{\texttt{PS}\xspace}	&84.91	&74.20	&\bd{93.20}	&61.64	&98.96	&\bd{2.300}	&\bd{1.368}	&\bd{2.487}	\\
\cline{2-11}											
	&	&softDice	&86.56	&73.93	&92.66	&61.49	&98.95	&2.568	&1.341	&2.760	\\
\rowcolor{mygray}\cellcolor{white}{}&\cellcolor{white}{ResUNet}
		&\ps	&\bd{95.93}	&\bd{82.44}	&92.15	&\bd{74.58}	&\bd{99.12}	&2.495	&1.451	&2.641	\\
\rowcolor{mygray}\cellcolor{white}{}&\cellcolor{white}{}
		&softDice+{\texttt{PS}\xspace}	&93.08	&77.55	&\bd{93.24}	&66.39	&99.04	&\bd{1.964}	&\bd{1.199}	&\bd{2.152}	\\
\bottomrule
\end{tabular}
}
\scalebox{0.465}{
\begin{tabular}{c | l | l | c c c c c | c c c}
\toprule
\mrtwo{Dataset}	&\mrtwo{Backbone}	&\mrtwo{Method}	& \multicolumn{5}{c|}{Volumetric Scores (\%) $\uparrow$}& \multicolumn{3}{c}{Topological Error $\downarrow$} \\
\cline{4-11}
&&	&AUC	&Dice	&Prec	&Recall	&ACC	&$\beta_0$	&$\beta_1$	&$\chi_{error}$	\\
\midrule											
	&	&softDice	&95.09	&66.24	&72.36	&61.08	&99.06	&2.225	&2.589	&2.919	\\
\rowcolor{mygray}\cellcolor{white}{}&\cellcolor{white}{UNet}
		&\ps	&93.59	&\bd{69.63}	&70.28	&\bd{68.99}	&99.01	&5.955	&2.244	&6.135	\\
\rowcolor{mygray}\cellcolor{white}{}&\cellcolor{white}{}
		&softDice+{\texttt{PS}\xspace}	&\bd{96.23}	&67.60	&\bd{73.74}	&62.40	&\bd{99.09}	&\bd{1.990}	&\bd{2.438}	&\bd{2.683}	\\
\cline{2-11}											
	&	&softDice	&95.07	&66.01	&71.94	&60.99	&99.05	&\bd{2.315}	&2.699	&\bd{3.011}	\\
\rowcolor{mygray}\cellcolor{white}{MassRoads}&\cellcolor{white}{UNet++}
		&\ps	&\bd{96.54}	&\bd{69.93}	&71.18	&\bd{68.73}	&99.03	&2.353	&\bd{2.407}	&3.033	\\
\rowcolor{mygray}\cellcolor{white}{}&\cellcolor{white}{}
		&softDice+{\texttt{PS}\xspace}	&96.05	&67.76	&\bd{73.06}	&63.17	&\bd{99.08}	&2.329	&2.582	&3.030	\\
\cline{2-11}											
	&	&softDice	&94.84	&65.84	&71.52	&61.01	&99.04	&2.816	&2.784	&3.512	\\
\rowcolor{mygray}\cellcolor{white}{}&\cellcolor{white}{ResUNet}
		&\ps	&95.42	&67.45	&70.83	&\bd{64.38}	&99.02	&5.666	&2.651	&5.409	\\
\rowcolor{mygray}\cellcolor{white}{}&\cellcolor{white}{}
		&softDice+{\texttt{PS}\xspace}	&\bd{96.43}	&\bd{67.52}	&\bd{73.00}	&62.82	&\bd{99.08}	&\bd{2.295}	&\bd{2.628}	&\bd{2.996}	\\
\midrule											
	&	&softDice	&96.59	&56.58	&62.56	&51.64	&99.60	&1.823	&1.157	&1.845	\\
\rowcolor{mygray}\cellcolor{white}{}&\cellcolor{white}{UNet}
		&\ps	&\bd{97.84}	&\bd{62.78}	&66.38	&\bd{59.55}	&99.62	&2.127	&1.059	&2.145	\\
\rowcolor{mygray}\cellcolor{white}{}&\cellcolor{white}{}
		&softDice+{\texttt{PS}\xspace}	&97.59	&61.17	&\bd{67.08}	&56.21	&\bd{99.63}	&\bd{1.358}	&\bd{1.001}	&\bd{1.380}	\\
\cline{2-11}											
	&	&softDice	&96.62	&57.42	&62.82	&52.87	&99.60	&1.481	&1.108	&1.503	\\
\rowcolor{mygray}\cellcolor{white}{DeepGlobe}&\cellcolor{white}{LinkNet34}
		&\ps	&95.04	&59.36	&63.85	&55.46	&99.60	&9.880	&1.933	&8.516	\\
\rowcolor{mygray}\cellcolor{white}{}&\cellcolor{white}{}
		&softDice+{\texttt{PS}\xspace}	&\bd{97.44}	&\bd{61.55}	&\bd{66.97}	&\bd{56.95}	&\bd{99.63}	&\bd{1.395}	&\bd{1.022}	&\bd{1.417}	\\
\cline{2-11}											
	&	&softDice	&96.08	&56.42	&62.37	&51.51	&99.60	&1.528	&1.121	&1.549	\\
\rowcolor{mygray}\cellcolor{white}{}&\cellcolor{white}{DinkNet34}
		&\ps	&95.54	&60.89	&64.20	&\bd{57.91}	&99.60	&7.341	&1.375	&6.520	\\
\rowcolor{mygray}\cellcolor{white}{}&\cellcolor{white}{}
		&softDice+{\texttt{PS}\xspace}	&\bd{97.71}	&\bd{61.71}	&\bd{66.89}	&57.27	&\bd{99.63}	&\bd{1.436}	&\bd{0.997}	&\bd{1.458}	\\
\bottomrule
\end{tabular}
}
\end{table}

\subsection{Main results}

Our \pointscatter can be regarded as an alternative for the segmentation approach. We compare our \pointscatter with two very competitive segmentation methods (\ie softDice~\cite{milletari2016v} and clDice~\cite{shit2021cldice}) on various mainstream backbone networks~\cite{ronneberger2015u,zhou2018unet++,zhang2018road,chaurasia2017linknet,zhou2018d}. 
We use the same training settings for the segmentation methods with our \ps, including the optimizer, training schedule, \etc. These methods are also implemented by MMSegmentation for a fair comparison. Except for using \ps directly, we also study the effect of using our \ps as an auxiliary task for the segmentation method. We combine these two methods as shown in Fig.~\ref{fig:unet-seg-ps} and use the sum of the loss function of these two methods as the objective to train the network. We denote this method as \psaux (abbreviation for \ps AUXiliary). We use the centerline labels to train the \ps branch in \psaux.

\subsubsection{Tubular Structure Segmentation. }
We exhibit the results in Table~\ref{tab:seg}. According to the volumetric metrics, we can conclude that our \ps achieves superior performance compared to the segmentation methods on most of the combinations of the datasets and the backbone networks, which confirms the effectiveness of our \ps. When applying \psaux to the segmentation method, we also observe improvements for most of the cases. The performance of \psaux certifies that the point set representation leads to better feature learning for the backbone network.
Our \ps obtains inferior performance than clDice according to the topology-based scores. We argue that it is because our \ps can capture more fine-scale structures which cannot be discovered by the segmentation models. These detailed predictions are beneficial to the volumetric scores but harmful to the topology. We will qualitatively analyze this phenomenon later. In addition, it is worth mentioning that \psaux can improve the topology scores as shown in Table~\ref{tab:seg}.

\subsubsection{Centerline Extraction. }
We also conduct extensive experiments to validate the advantage of \ps for the centerline extraction task. As shown in Table~\ref{tab:cline}, our \ps consistently surpasses the performance of the segmentation methods by a large margin according to the volumetric scores. Our \ps achieves similar precision to softDice, while complies with significantly higher recall values. It confirms again that our \ps can capture fine-scale details which cannot be detected by the segmentation model. The effect of \psaux is similar to the tubular structure segmentation task.

\subsection{Ablation study}

\subsubsection{Number of Points ($N$).}
We ablate the number of predicted points ($N$) within each scatter region in Table~\ref{tab:ablation_N}. With $D=4$, the maximum number of ground-truth points in each scatter region is 16. Therefore, the performance is not satisfactory when $N=8$. Increasing $N$ has marginal improvement on the performance when $N \ge 16$.

\subsubsection{Downsample rate ($D$).}
We compare the effect of different downsample rates $D$ in Table~\ref{tab:ablation_D}. For the DRIVE dataset, $D=4$ shows the best performance on the segmentation task while $D=2$ is slightly better on the centerline extraction task. For the MassRoads dataset, different $D$ yield similar performances on both tasks.

\subsubsection{Greedy Bipartite Matching.}
Our greedy bipartite matching is theoretically faster than the Hungarian method and can be easily implemented on GPU. We compare the running time in each training iteration of these two methods in Table~\ref{tab:speed}. We execute the greedy method on GPU TITAN RTX and the Hungarian algorithm on 
Intel(R) Xeon(R) CPU E5-2680 v4. The results show that our greedy method is at least three orders of magnitude faster than the Hungarian algorithm. The latency of our greedy method is negligible compared to the computation time of neural networks, whereas the latency of the Hungarian algorithm is unaffordable for large images.

\begin{table}[t]
\begin{minipage}[b]{.44\linewidth}
\centering
\caption{Ablation on $N$. ($D=4$)}
\label{tab:ablation_N}
\scalebox{0.6}{
\begin{tabular}{c | c | c c c | c c}
\toprule
\multicolumn{2}{c|}{} &\multicolumn{3}{c|}{Segmentation} &\multicolumn{2}{c}{Centerline}\\
\midrule
Dataset	& $N$	&Dice(\%)	&clDice(\%) &ACC(\%) &Dice(\%)	&ACC(\%) \\
\midrule											
\mr{4}{DRIVE}	&8	&64.80   &66.71   &92.27	&78.31  &97.79	\\
	            &16	&\bd{81.63}	&\bd{82.89}	&\bd{95.23}	&\bd{81.92}	&\bd{98.05}	\\
	            &32	&78.73   &80.73   &94.60	&79.07	&97.85	\\
	            &64	&78.33   &80.57   &94.54	&80.08	&97.91	\\
\midrule							
\mr{4}{MassRoads}	&8	&57.61       &58.23    &95.20   &64.73	&98.94	\\
	                &16	&\bd{77.57}  &86.42	   &96.87	&69.63	&99.01	\\
	                &32	&77.52   &\bd{86.55}   &96.87	&70.05	&99.03	\\
	                &64	&77.54   &86.40   &\bd{96.89}	&\bd{70.38}	&\bd{99.04}	\\
\bottomrule
\end{tabular}
}
\end{minipage}
\begin{minipage}[b]{.56\linewidth}
\centering
\caption{Ablation on $D$.}
\label{tab:ablation_D}
\scalebox{0.72}{
\begin{tabular}{c | c | c c c | c c}
\toprule
\multicolumn{2}{c|}{} &\multicolumn{3}{c|}{Segmentation} &\multicolumn{2}{c}{Centerline}\\
\midrule
Dataset	& $D$	&Dice(\%)	&clDice(\%) &ACC(\%) &Dice(\%)	&ACC(\%) \\
\midrule											
	    &2	&81.26   &82.16   &95.20	&\bd{82.70}	&\bd{98.07}	\\
DRIVE	&4	&\bd{81.63}	&\bd{82.89}	&\bd{95.23}	&81.92	&98.05	\\
	    &8	&79.80   &80.48   &94.77	&78.59	&97.81	\\
\midrule
	        &2	&\bd{77.90}  &\bd{86.92}  &\bd{96.93}	&\bd{69.87}	&99.02	\\
MassRoads	&4	&77.57  &86.42	&96.87	&69.63	&99.01	\\
            &8	&77.54  &86.31  &96.86	&68.79	&\bd{99.03}	\\
\bottomrule
\end{tabular}
}
\end{minipage}

\end{table}

\begin{table}
\begin{center}
\caption{Running time (seconds) of Greedy and Hungarian bipartite matching. We set $D=4$ and $\text{batchsize}=4$.}
\label{tab:speed}
\scalebox{0.8}{
\begin{tabular}{c|c|c|c}
\toprule
\textbf{Method} & Complexity & Image Size & Running Time (seconds)  \\
\hline
        &           & 384 $\times$ 384    & 0.0076 \\
Greedy  & $O(M^2)$  & 768 $\times$ 768    & 0.0100 \\
        &           & 1024 $\times$ 1024  & 0.0123\\
        \midrule
          &         & 384 $\times$ 384    & 3.0043 \\
Hungarian & $O(M^3)$ & 768 $\times$ 768    & 12.7027 \\
          &         & 1024 $\times$ 1024  & 20.9922\\
\bottomrule
\end{tabular}
}
\end{center}
\end{table}

\subsection{Qualitative Analysis}

\begin{figure}
\centering
\includegraphics[width=0.95\textwidth]{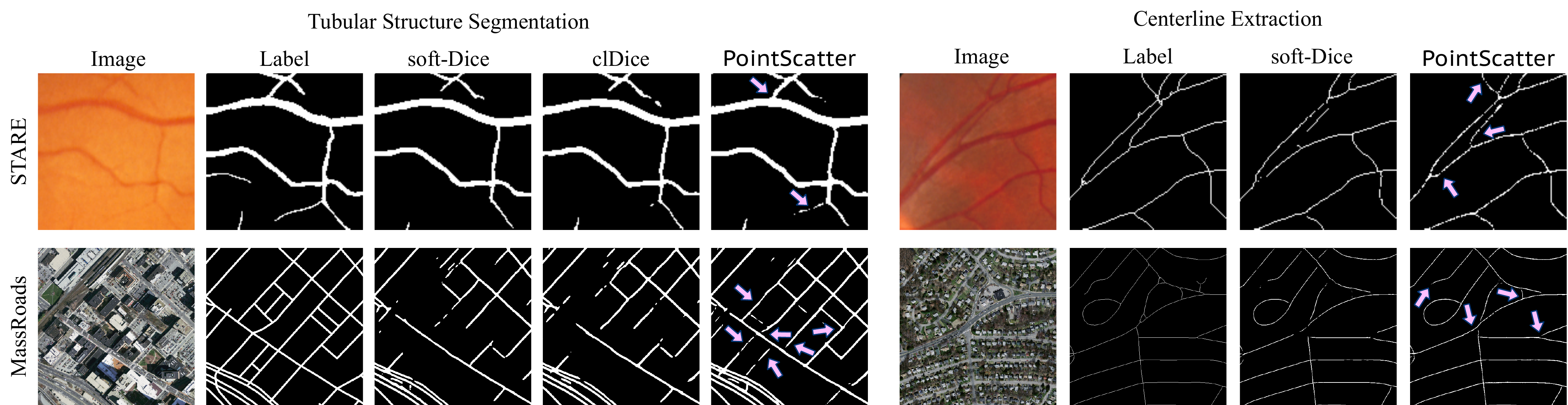}
\caption{Visual comparison for our \ps with other methods (zoom for details). The areas pointed by the arrows are missed by other models, while extracted by our \ps. More qualitative results can be found in the supplementary materials. }
\label{fig:vis}
\end{figure}

We qualitatively compare our method and the mask segmentation methods in Fig.~\ref{fig:vis}. Our \ps performs better on small branches or bifurcation points. 
It shows a better ability for our \ps to learn the complicated fine-scale information, which is contributed by the flexibility of the point set representation.
Note that sometimes the small branches detected by \ps are not densely connected (\eg the top left image), which decreases the performance on the topology-based metrics. However, it is better to extract tubular segments than miss the whole branch. We will leave future work to improve the topology performance of our \ps.

\section{Conclusion}

This paper proposes \ps, a novel architecture that introduces the point set representation for tubular structure extraction. This network can be trained end-to-end and efficiently with our proposed greedy bipartite matching algorithm. The extensive experiments reveal that our \ps achieves superior performance to the segmentation counterparts on the tubular structure segmentation task in most of the experiments, and significantly surpasses other methods on the centerline extraction task.

This novel design presents the potential of point set representation for tubular structures, and future work may include:
\begin{itemize}
\item Exploring the performance of \ps on the more challenging 3D tubular extraction tasks such as coronary vessel extraction.
\item Improving the topology of predicted points of \ps to enhance the performance of the topology-based metrics.
\item Promoting the point set representation for the general segmentation task.
\end{itemize}

\textbf{Acknowledgement.} This work is supported by Exploratory Research Project of Zhejiang Lab (No. 2022RC0AN02), Project 2020BD006 supported by PKUBaidu Fund.

\appendix
\newpage
\section{Appendix}

\captionsetup{font=footnotesize, labelfont=footnotesize}

\subsection{Implementation details}

In this section, we supplement the implementation details that are not covered in Section 4.1.

In Equation (3), the weight of regression loss is set to $\lambda=10$. As for the focal loss, we use $\alpha=0.6$ for the segmentation task and $\alpha=0.7$ for the centerline extraction task. The $\gamma$ in focal loss is set to $2.0$ for all circumstances. Specially, we set $\alpha=0.8$ for the centerline extraction task of MassRoads and DeepGlobe.
For all datasets except DeepGlobe, we use the ADAM optimizer with the initial learning rate 1e-3 and cosine learning rate schedule to train the network end-to-end. We use the poly learning rate schedule for Deepglobe and set the initial learning rate and the power to 1e-3 and 3, respectively.
The weight decay is set to be 1e-4 uniformly.

We then introduce the dataset specified hyper-parameters. Since the numbers of images are significantly different for the datasets, we set the number of iterations separately for each dataset. For DRIVE and STARE, we train the \ps for 3K iterations, and 10K for MassRoads. We use batchsize=4 for these three datasets. The DeepGlobe is much harder than the other datasets, hence we set its iteration number to 40K with batchsize=16. The sizes of input images are $384 \times 384$ for DRIVE and STARE, $1024 \times 1024$ for MassRoads and $768 \times 768$ for DeepGlobe.

\subsection{Ablation of the Matching Methods}

We compare the performance of the greedy bipartite matching method and the Hungarian algorithm on the two tasks in Table~\ref{tab:ablation_BMM_Seg} and Table~\ref{tab:ablation_BMM_Cline}. We use the same settings to train on these two methods for each dataset to achieve a fair comparison.
On both tasks, these two methods have similar volumetric scores, while the greedy method yields a slightly better topology-based score.

A succinct implementation of the greedy bipartite matching is demonstrated in Listing~\ref{listing:greedy_code}. Batching on all regions makes our implementation efficient.

\begin{listing}[ht]
\begin{minted}{python}
import torch

def batched_greedy_assignment(cost):
    """
    Args: 
        cost (Tensor): shape (num_region, num_pred (N), num_gt (K))
    Returns:
        assignment (LongTensor): shape (num_region, num_gt (K))
    """
    num_region, num_pred, num_gt = cost.shape 
    assignment = - torch.ones(
        size=(num_region, num_gt), 
        dtype=torch.long, 
        device=cost.device
    )

    for i in range(num_gt):
        cur_min_idx = torch.argmin(cost[..., i], dim=-1)
        assignment[..., i] = cur_min_idx
        index = cur_min_idx.view(num_region, 1, 1).expand(-1, -1, num_gt)
        cost.scatter_(dim=1, index=index, value=float('inf'))

    return assignment
\end{minted}
\caption{PyTorch codes of batched greedy bipartite assignment. This function assigns each ground-truth point to a predicted point uniquely. We assume that $K \le N$ in input cost tensor.}
\label{listing:greedy_code}
\end{listing}

\begin{table}
\centering
\caption{Comparison of the greedy and Hungarian matching algorithm on tubular structure segmentation task.}
\label{tab:ablation_BMM_Seg}
\scalebox{0.88}{
\begin{tabular}{c | c | c c c c | c c c}
\toprule
Dataset	&Method	&AUC(\%) &Dice(\%)	&clDice(\%) &ACC(\%) &$\beta_0$Error	&$\beta_1$Error	&$\chi_{error}$\\
\midrule											
\mr{2}{STARE}	&Greedy	    &97.86   &82.73   &\bd{85.83}   &97.45	&\bd{0.818}   &\bd{0.774}   &\bd{0.978}   \\
	            &Hungarian	&\bd{97.88}   &\bd{82.84}   &85.77   &\bd{97.48}	&0.861   &0.789   &1.021   \\
\midrule							
\mr{2}{MassRoads}	&Greedy	    &\bd{97.65}   &77.57   &86.42   &96.87	&0.944   &1.353   &1.616   \\
	                &Hungarian	&97.60   &\bd{77.65}   &\bd{86.60}   &\bd{96.88}	&\bd{0.902}   &\bd{1.323}   &\bd{1.558}  \\
\bottomrule
\end{tabular}
}
\end{table}

\begin{table}
\centering
\caption{Comparison of the greedy and Hungarian matching algorithm on centerline extraction task.}
\label{tab:ablation_BMM_Cline}
\scalebox{0.8}{
\begin{tabular}{c | c | c c c c c | c c c}
\toprule
Dataset	& Method	&AUC(\%) &Dice(\%)  &Prec(\%)   &Recall(\%) &ACC(\%) &$\beta_0$Error	&$\beta_1$Error	&$\chi_{error}$\\
\midrule											
\mr{2}{STARE}	&Greedy	    &\bd{94.52}   &\bd{81.77}   &\bd{92.09}   &\bd{73.52}   &\bd{99.10}	&\bd{2.158}   &\bd{1.424}   &\bd{2.303}   \\
	            &Hungarian	&93.95   &80.67   &91.85   &71.91   &99.07	&3.286   &1.702   &3.431   \\
\midrule							
\mr{2}{MassRoads}	&Greedy	    &93.59   &69.63   &70.28   &\bd{68.99}   &99.01	&5.955   &2.244   &6.135   \\
	                &Hungarian	&\bd{95.51}   &\bd{69.64}   &\bd{70.86}   &68.45   &\bd{99.02}	&\bd{4.315}   &\bd{2.288}   &\bd{4.766}  \\
\bottomrule
\end{tabular}
}
\end{table}

\clearpage
\subsection{Comparison with Other Methods}

In Table~\ref{tab:sota}, we further compare our method with previous approaches on the STARE dataset. We adopt the data split method in RV-GAN~\cite{kamran2021rv} for fair comparison. Our methods achieve SOTA performance on AUC and clDice, and achieve competitive Dice and ACC scores compared with other methods.

\begin{table}
	\centering
	\caption{Comparison with previous works on the vessel segmentation task. We follow the data split methods in RV-GAN. Our models are based on the U-Net backbone.}
	\label{tab:sota}
    \renewcommand\arraystretch{1.2}
    \begin{tabular}{c | l | c c c c}
        \toprule
        Dataset	&Method	&AUC(\%)	&Dice(\%)	&clDice(\%)	&ACC(\%)		\\
        \midrule
        &R2UNet~\cite{alom2018recurrent}	&99.14   &\bd{84.75}   &-   &97.12   	\\
        &CE-Net~\cite{gu2019net}	&98.71   &83.13   &85.87   &97.59   \\
        &DUNet~\cite{jin2019dunet}	&98.32   &81.43   &-   &96.41   \\
        STARE&IterNet~\cite{li2020iternet}	&99.15   &81.46   &-   &\bd{97.82}  \\
        &RV-GAN~\cite{kamran2021rv} &98.87   &83.23   &-   &97.54 \\
        \rowcolor{mygray}\cellcolor{white}{}
        &\ps	&\bf{99.24}   &84.52   &87.46   &97.75		\\
        \rowcolor{mygray}\cellcolor{white}{}
        &softDice+{\psaux} &98.59   &84.51   &87.36   &97.75	\\
        \rowcolor{mygray}\cellcolor{white}{}
        &clDice+{\psaux} &98.59   &84.74  &\bd{88.29}   &97.77	\\
        \bottomrule
    \end{tabular}
\end{table}

\subsection{Additional Qualitative Results}

\begin{figure}
\centering
\includegraphics[width=1.0\textwidth]{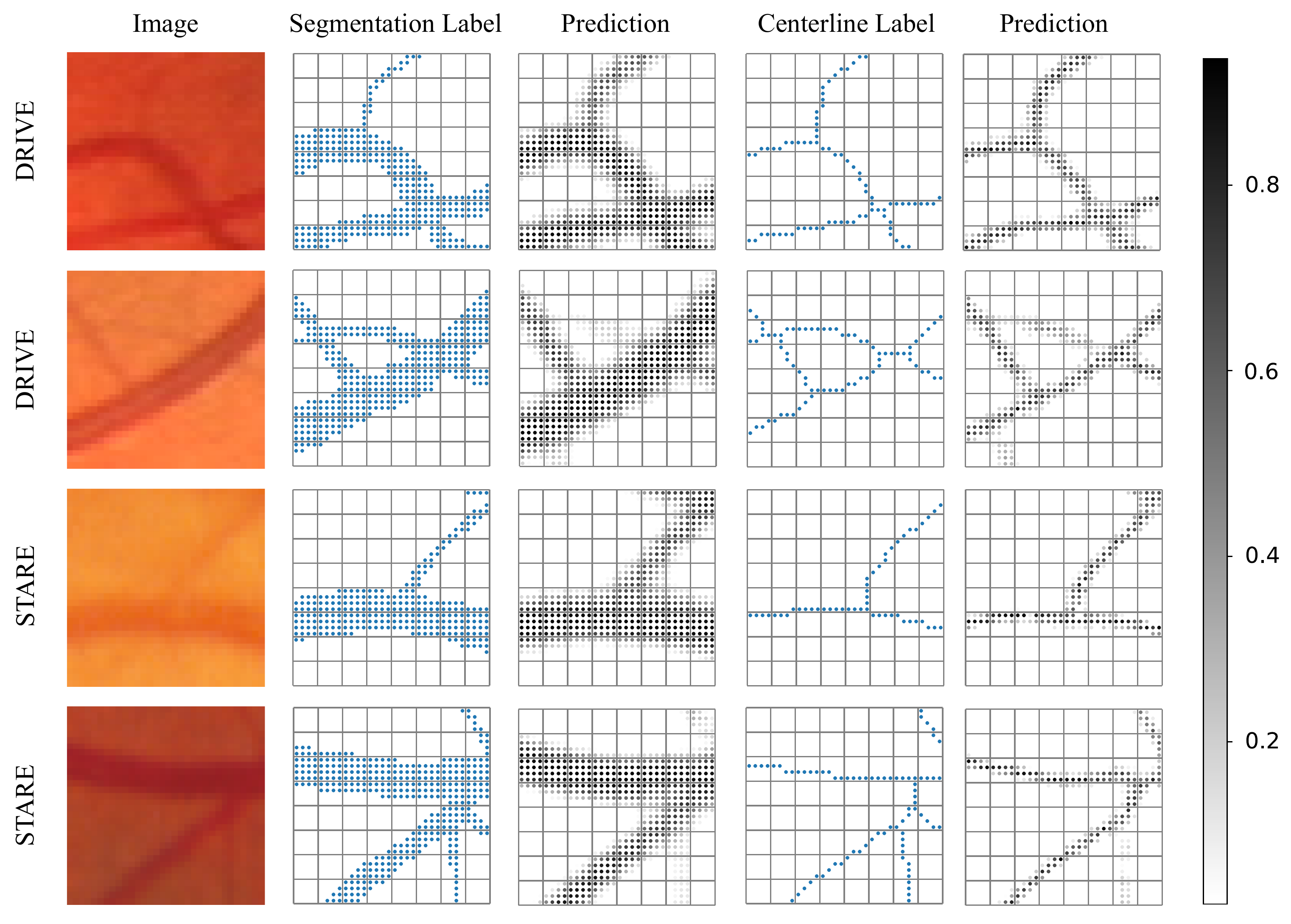}
\caption{Illustration of the points labels and points predictions of our \ps on the two tasks. We set $D=4$ in our experiments and the each $4 \times 4$ bin in the images is corresponding to a scatter region.}
\label{fig:supp_point}
\end{figure}

In this section, we provide more qualitative analysis of our \ps. In Fig.~\ref{fig:supp_point}, we show the points labels and points predictions of our \ps. We use the gray value to represent the objectness score of each predicted point, the darker, the higher. 
Since the GT points are tiled as grid shape, the predicted points are also located in the center of each pixel. 

\begin{figure}
\centering
\includegraphics[width=1.0\textwidth]{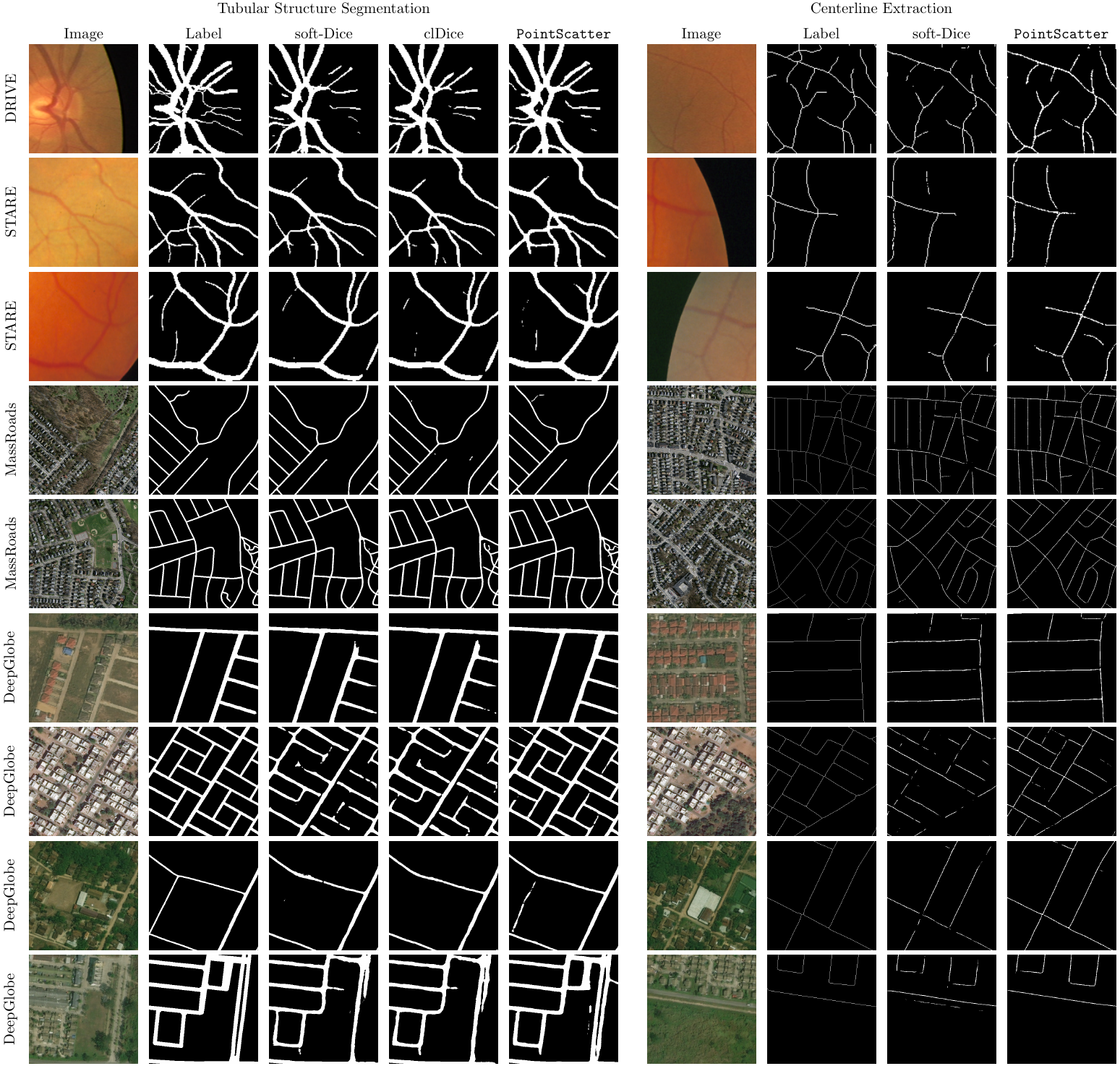}
\caption{Additional visual comparison for our \ps with other methods.}
\label{fig:add_visual}
\end{figure}

We also illustrate more tubular segmentation and centerline extraction results in Fig.~\ref{fig:add_visual}.

\clearpage
%
%
\bibliographystyle{splncs04}
\bibliography{egbib}

\end{document}